\pdfoutput=1
\documentclass[11pt]{article}

\usepackage[preprint]{acl}

\usepackage{times}
\usepackage{latexsym}
\usepackage{amsmath}
\usepackage{booktabs}
\usepackage[T1]{fontenc}
\usepackage[utf8]{inputenc}

\usepackage{microtype}

\usepackage{inconsolata}

\usepackage{graphicx}

\title{Trajectory-Guided Steering: Mitigating Spurious Feature Bias in Clinical Classification}

\author{
  \textbf{Fardin Ahsan Sakib}$^{1}$,
  \textbf{Ziwei Zhu}$^{1}$,
  \textbf{Özlem Uzuner}$^{2}$
\\
  $^{1}$Department of Computer Science, $^{2}$Department of Information Sciences and Technology \\
  George Mason University \\
  \texttt{\{fsakib,zzhu20,ouzuner\}@gmu.edu}
}
\begin{document}
\maketitle
\begin{abstract}
Large language models rely on spurious features in clinical classification: swapping a single gender token in clinical notes alters gender-irrelevant diagnoses in up to 53.5\% of cases. Existing inference-time corrections either need counterfactual data built around the specific spurious feature or behave unpredictably. We observe that how a model's preference between answers evolves across layers, its prediction trajectory, indicates for each document whether a correction is likely to help. We propose Trajectory-Guided Steering (TGS), which decides from the trajectory alone whether to correct each prediction and in which direction, with no counterfactual data and no knowledge of the spurious feature. On 16 clinical classification tasks from i2b2, TGS fixes 17.7\% of errors while breaking only 1.3\% of correct predictions and stays net-positive across all tested hyperparameter settings; all baselines are net-negative or near zero. The same corrections reduce reliance on the spurious feature: on the most affected diseases, gender-swap inconsistency falls sharply (hypertension 53.5\%$\rightarrow$38.0\%, diabetes 35.1\%$\rightarrow$21.5\%). Applying the correction to every document instead raises breaks on correct positive predictions from 0.1\% to 6.4\%, so per-document selection is essential. TGS also transfers to three model families and three other clinical and biomedical datasets.
\end{abstract}
\section{Introduction}
\label{sec:intro}

Large language models are increasingly applied to clinical NLP tasks such as screening, triage, and diagnostic support. Yet their predictions can depend on spurious features, aspects of the input irrelevant to the clinical question. This is documented in clinical use: changing only a patient's gender or race affects GPT-4's ability to rank the correct diagnosis first in 37\% of cases \citep{zack2024assessing}, LLMs repeat race-based medical misconceptions \citep{omiye2023large}, and they infer drug use from mentions of alcohol or smoking \citep{sakib2025-sdoh}. Consider a discharge summary documenting Type~2 diabetes with multiple insulin prescriptions: with identical symptoms and medications, the model's predicted diagnosis flips from positive to negative when only the gender word changes, even though gender carries no diagnostic information for this disease. \textbf{Such reliance makes the model's predictions untrustworthy: they can change with information that has no bearing on the diagnosis.} To quantify it, we swap a single gender token across 16 diseases for which gender is diagnostically irrelevant, holding all clinical evidence constant. Because the evidence alone determines the correct answer, any change in prediction is an error; this \emph{inconsistency} occurs in \textbf{9.7\% to 53.5\% of cases} (Appendix~\ref{sec:appendix_inconsistency}).

\begin{figure*}[t]
    \centering
    \includegraphics[width=\textwidth]{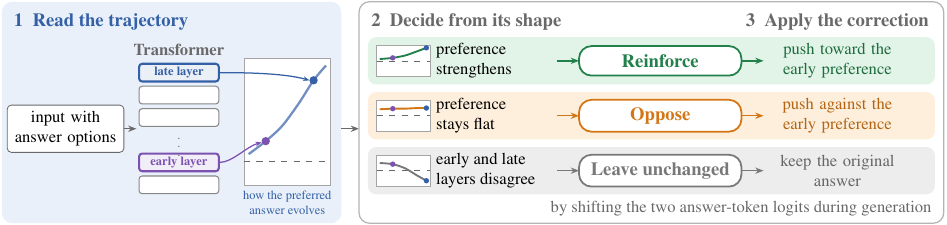}
    \caption{Overview of Trajectory-Guided Steering (TGS). Before generating, TGS reads which answer the model prefers at each layer and how strongly, giving a prediction trajectory. Comparing an early and a late layer, it finds whether the preference strengthens, stays flat, or changes between the two. Based on this shape, TGS pushes toward the early preference, pushes against it, or keeps the original answer, by shifting the two answer-token logits during generation.}
    \label{fig:method}
\end{figure*}

Existing ways to reduce reliance on spurious features are costly. Training-time methods such as invariant or domain-adversarial learning require retraining, often with labels for the feature or for distinct environments \citep{arjovsky2020invariantriskminimization, ganin2016domainadversarialtrainingneuralnetworks}. Inference-time methods avoid retraining, but prompt changes have unpredictable effects, and activation steering (e.g., CAA \citep{rimsky-etal-2024-steering}, FairSteer \citep{li2025-fairsteer}) or editing demographic information inside the model \citep{ahsan2025elucidating} needs counterfactual training pairs or knowledge of which feature drives the errors.

Clinical settings rarely provide either. Records are restricted by data use agreements \citep{agrawal2022largelanguagemodelsfewshot}, counterfactual data requires expert annotation, and the features that sway a model are rarely known in advance. A correction must also be selective, because changing a correct clinical prediction is itself a harm. \textbf{Clinical classification therefore needs a correction that works at inference time, uses no additional data, and leaves correct predictions intact.}

We propose Trajectory-Guided Steering (TGS), which decides from the model's own forward pass whether and how to correct each prediction. Using the logit lens \citep{nostalgebraist2020logitlens}, which projects each layer's hidden state through the unembedding matrix, TGS tracks the model's preference between the answer tokens across layers; we call this curve the \emph{prediction trajectory}. Its shape, whether early and late layers agree and whether the preference strengthens, indicates when a correction is safe and in which direction. When it is, a second forward pass adjusts the logits of the answer tokens toward the direction the trajectory supports; otherwise, the prediction is left unchanged. TGS needs no counterfactual data and no knowledge of the spurious feature. Empirically, these corrections also reduce the model's reliance on the gender token.

On the i2b2 clinical benchmark \citep{uzuner2009-obesity-challenge}, TGS fixes 17.7\% of prediction errors while breaking only 1.3\% of correct predictions. Its net gain exceeds that of every baseline. The same corrections reduce gender-swap inconsistency, most on the diseases where gender sways the model most (e.g., hypertension 53.5\%$\rightarrow$38.0\%). TGS also transfers to three model families and three other clinical and biomedical datasets. Figure~\ref{fig:method} provides an overview. Our contributions are threefold. (1)~\textbf{A new signal for correction}: we show that logit-lens trajectory shape distinguishes correctable errors from predictions that should be left untouched, turning a layer-wise readout of the model's predictions into a control signal. (2)~\textbf{Selective, per-document correction from the model's own forward pass}: unlike activation steering (CAA, FairSteer), TGS needs neither counterfactual data nor knowledge of the spurious feature. (3)~\textbf{Robust net-positive performance that baselines do not achieve}: TGS is net-positive across three model families and four clinical and biomedical datasets, and on i2b2 steering and prompting baselines are net-negative or near zero; an ablation shows that per-document selection is essential for preserving correct predictions.

\section{Prediction Trajectories Reveal Three Trajectory Shapes}
\label{sec:trajectory}

An inference-time correction method must decide, for each document, whether to correct and in which direction. Without counterfactual data or knowledge of which feature drives the errors, that decision has to come from the model's own forward pass. Our goal is to understand whether the model's layer-by-layer predictions differ systematically between correct and incorrect outputs, and if so, whether these differences could indicate when correction is likely to help.

\subsection{Prediction Trajectories}
We analyze LLaMA 3.1 8B Instruct \citep{llama3.1} on the i2b2 2008 Obesity Challenge \citep{uzuner2009-obesity-challenge}, a clinical benchmark where models classify each of 16 diseases as present or absent in discharge summaries. At every layer, we use the logit lens \citep{nostalgebraist2020logitlens} to project the hidden state through the model's unembedding matrix and obtain logit values for each answer token (e.g., \texttt{Yes} and \texttt{No}). The difference between these logits, the \emph{logit margin}, shows which answer the model favors at that depth; its size, which we call \emph{conviction}, shows how strongly. Tracking the margin from early to late layers produces a per-document \emph{prediction trajectory}: a curve showing whether the margin grows, stalls, or changes sign across depth. We orient each trajectory so that the model's preferred answer points upward, letting documents that favor opposite answers be compared by the shape of their trajectory.

\subsection{Three Trajectory Shapes}
\begin{figure*}[t]
    \centering
    \includegraphics[width=\textwidth]{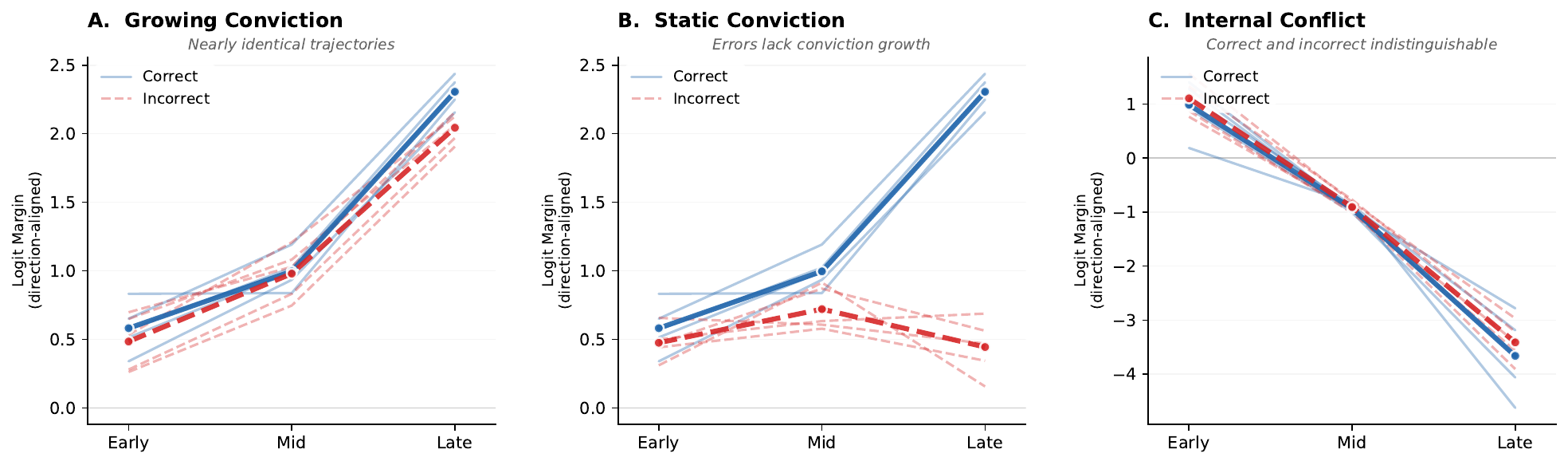}
    \caption{Three trajectory shapes on i2b2 (LLaMA 3.1 8B), each implying a different safe action. The y-axis shows the direction-aligned logit margin: the difference between the logit values of the two answer tokens, oriented so that the model's preferred direction points upward. Thin lines: individual documents; thick line with markers: group mean. Blue solid: correct predictions; red dashed: incorrect predictions. \textbf{Panel~A}: Growing conviction (reinforce). Correct and incorrect follow nearly identical trajectories. \textbf{Panel~B}: Static conviction (oppose). Incorrect predictions fail to accumulate the conviction growth seen in correct predictions. \textbf{Panel~C}: Internal conflict (leave unchanged). Correct and incorrect are fully indistinguishable.}
    \label{fig:patterns}
\end{figure*}

Examining these trajectories on i2b2, we find they naturally partition into three shapes (Figure~\ref{fig:patterns}): \textbf{growing conviction} (early and late layers agree and conviction grows), \textbf{static conviction} (they agree but conviction does not grow), and \textbf{internal conflict} (they disagree). \textbf{Each shape implies a distinct correction: reinforce the preferred answer, oppose it, or leave the prediction unchanged.} We motivate these below and formalize them in Section~\ref{sec:method}.

In the first shape, \emph{growing conviction} (Panel~A), early and late layers agree on the preferred answer and conviction grows substantially across layers. Correct and incorrect predictions follow nearly identical trajectories, so the trajectory alone cannot distinguish them. Yet correction remains possible: in most incorrect predictions of this shape, the trajectory favors the correct answer, but the generated answer differs. Reinforcing the preferred answer moves the output toward it, which recovers some of these errors while most correct predictions keep their answer.

In the second shape, \emph{static conviction} (Panel~B), early layers establish a preference but conviction growth is limited. Here, in contrast, the trajectory does distinguish correct from incorrect: correct predictions continue to accumulate conviction while incorrect predictions do not. A preference that deeper layers do not strengthen marks errors that could be safely opposed, while correct predictions, whose conviction grows, would not be targeted.

In the third shape, \emph{internal conflict} (Panel~C), early and late layers disagree: the margin changes sign across layers. Correct and incorrect predictions are fully indistinguishable in this shape: their trajectories overlap completely. But here no consistent direction exists for correction, so any correction is equally likely to harm a correct prediction as to fix an incorrect one. The safe action is to leave these documents unchanged.

Applying the same logit lens inspection to Qwen3~8B \citep{qwen3}, a different model family, reveals the same three qualitative shapes. The structural properties are preserved: growing-conviction trajectories show the same correct--incorrect overlap, and internal-conflict trajectories remain indistinguishable across outcomes. This suggests the shapes are not an artifact of a single architecture. TGS results for Qwen3 are reported in Section~\ref{sec:results}.

\subsection{From Shapes to a Correction Rule}
The three shapes yield a single rule, applied per document: reinforce the preferred answer when conviction grows, oppose the preference when conviction stays static, and leave the prediction unchanged otherwise, including when early and late layers conflict. \textbf{Safe correction does not require distinguishing correct from incorrect predictions.} Even where their trajectories are identical (growing conviction), reinforcing the shared direction fixes errors while leaving correct predictions largely intact, because it only strengthens a preference the trajectory already supports. Section~\ref{sec:method} formalizes these mappings into a single per-document correction signal.

%% ============================================================
%% SECTION 4: METHOD
%% ============================================================
\section{Method}
\label{sec:method}

\paragraph{Problem.} Consider an $L$-layer language model and a classification task whose $K$ candidate answer tokens are specified in the prompt. Our goal is a per-document correction of prediction errors at inference time, without counterfactual data or knowledge of the spurious feature, while leaving correct predictions intact. We formalize the binary case ($K{=}2$) with answer tokens $t^+$ and $t^-$; the quantities below generalize to $K > 2$ (Appendix~\ref{sec:appendix_multiclass}).

TGS is a training-free inference-time method that operates in two passes (Figure~\ref{fig:method}). Pass~1 extracts a per-document logit-lens trajectory and maps it to a scalar correction signal $e_d$ that encodes whether and how to correct (Sections~\ref{sec:trajectory_extraction}--\ref{sec:routing}). Pass~2 applies a logit bias to documents with $e_d \neq 0$ and leaves all others unchanged (Section~\ref{sec:generation}).

\subsection{Trajectory Extraction}
\label{sec:trajectory_extraction}

Given an input sequence, we perform a single forward pass through all $L$ layers of the model without generating any tokens; the answer itself is produced later, during generation (Section~\ref{sec:generation}). At each layer $\ell \in \{0, \ldots, L{-}1\}$, we project the final-token hidden state through the model's own final normalization layer and language model head:
\begin{equation}
\label{eq:logits}
\mathbf{z}_\ell = W_U \cdot \text{Norm}(\mathbf{h}_\ell^{[-1]})
\end{equation}
where $\mathbf{h}_\ell^{[-1]}$ is the hidden state at the last input position in layer $\ell$, and $W_U$ is the unembedding matrix. This is the \emph{logit lens} projection \citep{nostalgebraist2020logitlens}, which reveals the model's implicit token preferences at intermediate layers.

From the projected logits, we extract two signals at each layer. The \textbf{margin} captures the model's preference between the positive and negative answer tokens:
\begin{equation}
\label{eq:margin}
m_\ell = \mathbf{z}_\ell[t^+] - \mathbf{z}_\ell[t^-]
\end{equation}
where $t^+$ and $t^-$ are the answer tokens for the positive and negative classes (e.g., \texttt{Yes}/\texttt{No}), defined per task in Section~\ref{sec:setup} (prompts in Appendix~\ref{sec:appendix_prompts}). The \textbf{binary entropy} measures how sharply the model favors one answer over the other:
\begin{align}
H_\ell &= -\sum_{k \in \{t^+, t^-\}} p_k \log p_k \label{eq:entropy} \\
p_k &= \frac{\exp(\mathbf{z}_\ell[k])}{\exp(\mathbf{z}_\ell[t^+]) + \exp(\mathbf{z}_\ell[t^-])} \nonumber
\end{align}

\paragraph{Steering layer selection.}
Early logit-lens projections tend to be noisy and unreliable \citep{nostalgebraist2020logitlens, belrose2023eliciting}. We select the steering layer $\ell^*$ per document by identifying the first significant drop in binary entropy across intermediate layers, which marks where the model first commits to a decision. This makes $\ell^*$ adaptive: it varies across documents and models without manual tuning. The full selection procedure and cross-model analysis are provided in Appendix~\ref{sec:appendix_steering}.

\paragraph{Key signals.}
From the trajectory, we extract two anchor points:
\begin{equation}
\label{eq:signals}
m_{\mathrm{e}} = m_{\ell^*}, \quad m_{\mathrm{l}} = m_{L-1}
\end{equation}
where $m_{\mathrm{e}}$ is the margin at the steering layer and $m_{\mathrm{l}}$ the margin at the final layer. Together, these capture the model's early preference and how it evolves across depth.

\subsection{Trajectory-Guided Routing}
\label{sec:routing}

The trajectory signals from Pass~1 determine a per-document correction signal $e_d$ that captures both whether to correct and in which direction. The guiding principle is conservative: TGS modifies predictions only when the trajectory provides clear evidence for the direction of correction, and defaults to leaving the original prediction unchanged.

We now formalize the routing rule, which assigns each document to one of the three actions (reinforce, oppose, or leave unchanged) for the shapes identified in Section~\ref{sec:trajectory}.

Formally, let $s_\ell = \operatorname{sign}(m_\ell)$. If $s_{\ell^*} \neq s_{L-1}$, the trajectory shows internal conflict and $e_d = 0$. When they agree, we measure conviction growth $|m_{\mathrm{l}}| - |m_{\mathrm{e}}|$, the increase in conviction regardless of direction:
\begin{equation}
\label{eq:routing}
e_d = \begin{cases}
+m_{\mathrm{e}} & |m_{\mathrm{l}}| - |m_{\mathrm{e}}| > \tau \\
-\max(m_{\mathrm{e}},\; 0) & \text{otherwise}
\end{cases}
\end{equation}
where $\tau$ is the conviction-growth threshold. Intuitively, $e_d$ reinforces the early preference when conviction grew and opposes it when conviction stayed static with a positive early margin; all other documents, including static negative early margins, retain their original prediction ($e_d = 0$).

\subsection{Steered Generation}
\label{sec:generation}

For documents with $e_d \neq 0$, Pass~2 applies a logit bias during generation, adding $+\alpha e_d$ to the positive answer-token logit and $-\alpha e_d$ to the negative one, where $\alpha$ is the correction strength. The answer-token margin (Eq.~\ref{eq:margin}) therefore becomes:
\begin{equation}
\label{eq:bias}
m' = \big(\mathbf{z}[t^+] + \alpha e_d\big) - \big(\mathbf{z}[t^-] - \alpha e_d\big) = m + 2\alpha\, e_d .
\end{equation}
All other logits remain unchanged. For documents with $e_d = 0$, generation proceeds unmodified.

\subsection{Hyperparameters and Runtime}
\label{sec:hyperparams}

TGS has two hyperparameters. The correction strength $\alpha$ controls the magnitude of the logit bias and is set per model and dataset, reflecting differences in logit scale. The conviction-growth threshold $\tau$ separates growing from static conviction; it is robust across its tested range (Section~\ref{sec:analysis}). Both are selected via grid search, following the same evaluation protocol as the baselines; selecting them on a held-out 30\% split instead gives the same conclusions (Section~\ref{sec:analysis}). The sensitivity analysis and per-dataset grids are reported in Appendix~\ref{sec:appendix_grids}.

\paragraph{Runtime.} Trajectory extraction (Pass~1) adds one forward pass per document with no token generation. Documents with $e_d = 0$ incur only this extraction cost, while documents with $e_d \neq 0$ additionally undergo a modified generation pass.

TGS extends to $K > 2$ answer classes; Appendix~\ref{sec:appendix_multiclass} gives the formulation and an initial 3-class evaluation.

%% ============================================================
%% SECTION 5: EXPERIMENTAL SETUP
%% ============================================================
\section{Experimental Setup}
\label{sec:setup}

\subsection{Datasets}
\label{sec:datasets}

\paragraph{i2b2 Obesity Challenge (primary).}
We evaluate on the i2b2 2008 Obesity Challenge \citep{uzuner2009-obesity-challenge}, comprising clinical discharge summaries annotated for 16 disease conditions (e.g., diabetes, hypertension, asthma) as present or absent, with 243--598 documents per disease. Each document is prompted under both gender conditions, yielding two predictions per document; the note itself is unchanged.

\paragraph{n2c2 2018 Cohort Selection (second clinical benchmark).}
To test transfer to a new clinical dataset, we evaluate on the n2c2 2018 cohort selection task \citep{stubbs2019cohort}, in which 288 patients are annotated for 13 clinical-trial eligibility criteria as met or not met. We ask one Yes/No question per patient and criterion, using the i2b2 prompt with the criterion in place of the disease, under both gender conditions.

\paragraph{SHAC Drug Use (different spurious feature).}
To test a different spurious feature, we evaluate on drug use status classification from the Social History Annotation Corpus (SHAC) \citep{shac}, a social determinants of health corpus of MIMIC-III \citep{johnson2016mimic} clinical notes. Models determine whether drug use is present (current or past) or absent (none/unknown). Margins are computed via logsumexp over each token set to accommodate the multi-token outcome space ($t^+ = \{\text{current, past}\}$, $t^- = \{\text{none, not applicable}\}$).

\paragraph{PubMedQA (beyond clinical notes).}
As a first step beyond clinical notes, we evaluate on PubMedQA \citep{jin2019-pubmedqa}, a biomedical QA dataset where models determine whether published evidence supports a given research question. We use the 890 expert-annotated instances with a yes or no label.

\subsection{Models}
\label{sec:models}

All experiments use 7--8B parameter instruction-tuned LLMs loaded in bfloat16. Our primary model is LLaMA 3.1 8B Instruct \citep{llama3.1}. To test cross-model generalization, we additionally evaluate Mistral 7B Instruct v0.3 \citep{mistral} and Qwen3 8B \citep{qwen3}.

\subsection{Baselines}
\label{sec:baselines}

\paragraph{Prompt-based.} We compare against three prompt modification strategies that require no training data: (1) \textit{system-clinical}, a system prompt instructing the model to base assessments on clinical findings only; (2) \textit{system-objective}, a system prompt requesting objective, neutral clinical evaluation; and (3) \textit{chain-of-thought} (CoT) \citep{wei2022-cot}, which elicits step-by-step reasoning before the final answer.

\paragraph{Activation steering.} We compare against two activation steering methods. \textbf{CAA} \citep{rimsky-etal-2024-steering} computes a steering vector as the mean activation difference between contrastive prompt pairs, then adds it at a single selected layer across all response-region token positions. \textbf{FairSteer} \citep{li2025-fairsteer} trains a per-layer linear probe to detect biased activations and conditionally applies a debiasing steering vector only when the probe fires. Both require counterfactual training pairs; we reserve 30\% of documents for training. We report the best universal configuration for each, selected by sweeping hyperparameters on the test split.

\subsection{Evaluation}
\label{sec:evaluation}

Each document is classified by the outcome of its original prediction (TP, TN, FP, or FN), and the steered prediction determines whether the correction produces a \textit{fix} (incorrect $\rightarrow$ correct), a \textit{break} (correct $\rightarrow$ incorrect), or no change. We report fix rate (fixes / total errors), break rate (breaks / total correct predictions), and net gain (total fixes $-$ total breaks).

For i2b2 and n2c2 2018, each document yields two predictions (one per gender condition), counted independently for fix and break rates; gender-swap inconsistency is measured separately (Appendix~\ref{sec:appendix_inconsistency}). The i2b2 prompt includes a Questionable option; documents with Questionable ground truth have no definitive Yes/No label and are excluded (23 documents, ${<}$0.3\%). Original predictions that match none of the prompt's answer options (non-answers) are excluded from the counts, and steered non-answers are counted as incorrect. For SHAC and PubMedQA, each document yields one prediction. Our method evaluates on 100\% of documents; CAA and FairSteer evaluate on ${\sim}$70\% (30\% reserved for training). For a direct comparison, we additionally evaluate TGS on this same ${\sim}$70\% test subset (Table~\ref{tab:baselines}).

%% ============================================================
%% SECTION 6: RESULTS
%% ============================================================
\section{Results}
\label{sec:results}

\subsection{TGS Outperforms All Baselines}
\label{sec:primary-results}

\begin{figure*}[t]
    \centering
    \includegraphics[width=0.66\textwidth]{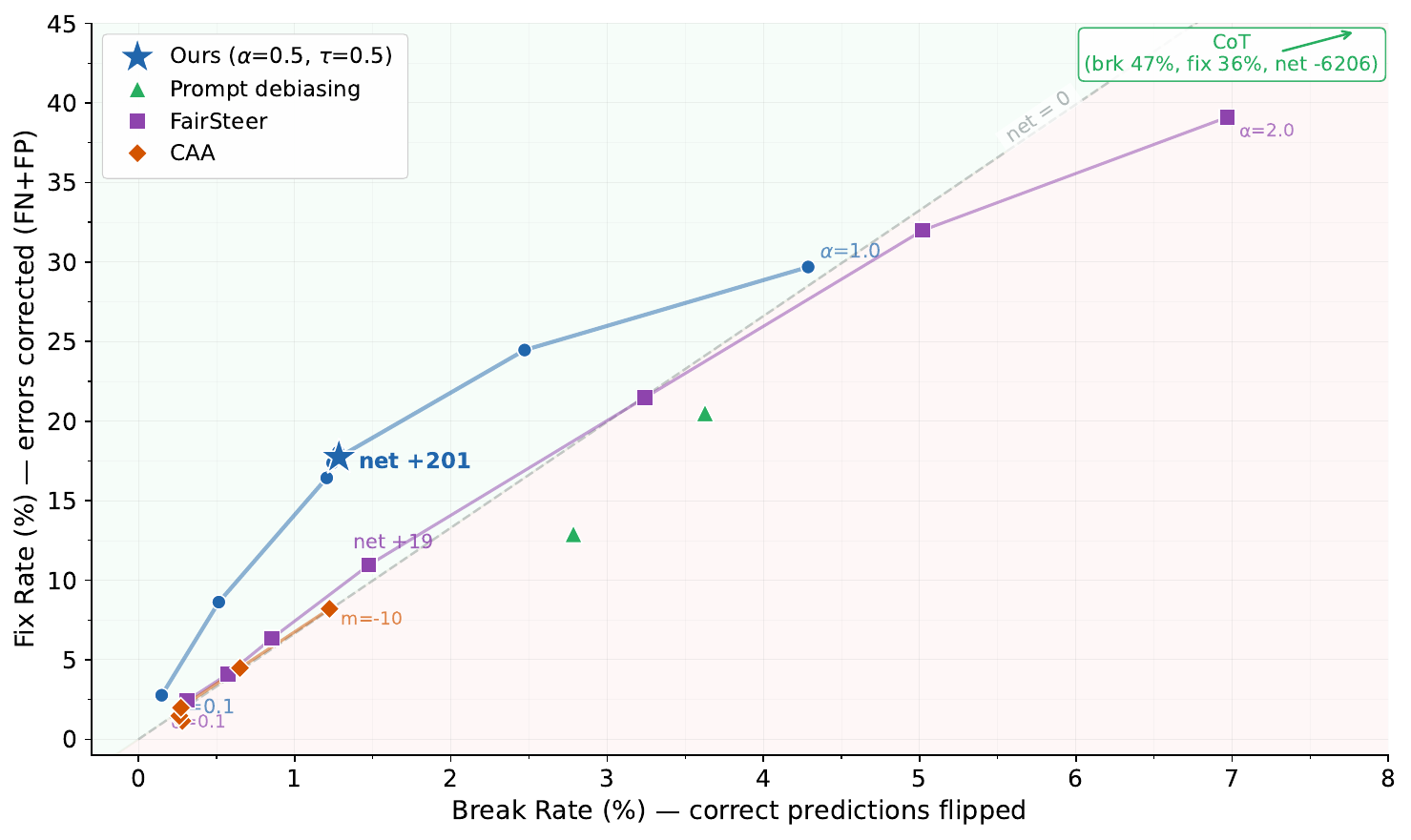}
    \caption{Fix rate vs.\ break rate tradeoff on i2b2 (LLaMA, 16 diseases). The green upper-left region is preferable: it combines high error correction (fix rate) with little damage to correct predictions (break rate); points above the dashed net-zero line are net-positive. TGS's fix--break curve (blue, varying $\alpha$ from 0.1 to 1.0) lies entirely in this region, and the starred point marks our reported configuration ($\alpha{=}0.5$, $\tau{=}0.5$). Baselines cluster near zero fix rate (CAA, labeled by its steering multiplier $m$) or cross into net-negative territory at higher strengths (FairSteer, prompts). CoT (off-chart) achieves 35.7\% fix rate but at 47.5\% break rate.}
    \label{fig:pareto}
\end{figure*}

\textbf{On i2b2, TGS achieves a far larger net gain than any baseline: it fixes 17.7\% of prediction errors (net~+201) while breaking just 1.3\% of correct predictions, whereas baselines are net-negative or only marginally positive} (Table~\ref{tab:baselines}). Figure~\ref{fig:pareto} makes this visible: TGS's fix--break curve (blue) stays inside the green low-break region as its correction strength rises from $\alpha{=}0.1$ to $1.0$, adding fixes gradually, whereas baseline points remain near zero or fall into the net-negative region at stronger settings.

Methods that achieve high fix rates do so at the cost of high break rates (Figure~\ref{fig:pareto}): CoT reaches 35.7\% fix but 47.5\% break. At their best configurations, CAA achieves net +3 and FairSteer net +19 on their ${\sim}$70\% test subset, compared with TGS's +146 on the same subset (Table~\ref{tab:baselines}). All three prompt strategies are net-negative, so among methods that need no counterfactual data, TGS is the only net-positive one.

\textbf{These corrections also reduce reliance on the spurious feature: gender-swap inconsistency drops from 25.1\% to 21.5\% across 16 diseases, with the largest reductions on the most affected diseases (hypertension 53.5\% $\rightarrow$ 38.0\%, diabetes 35.1\% $\rightarrow$ 21.5\%; Appendix~\ref{sec:appendix_inconsistency}).} Because each document is corrected from its trajectory alone, with no knowledge of the spurious feature, this reduction is empirical. Still, TGS fixes 19.8\% of errors that flip with the gender token, four times its rate on errors that persist under both genders (4.9\%). Adding the same logit bias toward the correct answer to every error corrects both kinds of error at nearly the same rate (13.7\% vs.\ 14.8\%), so the corrections of TGS concentrate on errors driven by the spurious feature (Appendix~\ref{sec:appendix_inconsistency}).

\begin{table}[t]
\centering
\small
\setlength{\tabcolsep}{4pt}
\begin{tabular}{lcrrr}
\toprule
Method & Train & Net\,$\uparrow$ & Fix\%\,$\uparrow$ & Break\%\,$\downarrow$ \\
\midrule
\multicolumn{5}{l}{\textit{Full test set}} \\
\textbf{TGS (ours)} & none & \textbf{+201} & 17.7 & 1.3 \\
Prompt: clinical & none & $-$126 & 12.9 & 2.8 \\
Prompt: objective & none & $-$83 & 20.5 & 3.6 \\
Prompt: CoT & none & $-$6206 & 35.7 & 47.5 \\
\midrule
\multicolumn{5}{l}{\textit{Same ${\sim}$70\% subset (CAA/FairSteer reserve 30\% to train)}} \\
\textbf{TGS (held-out)} & none & \textbf{+146} & 17.7 & 1.2 \\
CAA & 30\% & +3 & 2.0 & 0.3 \\
FairSteer & 30\% & +19 & 11.0 & 1.5 \\
\bottomrule
\end{tabular}
\caption{Baseline comparison on i2b2 (LLaMA, 16 diseases). Each block compares methods on identical data: the full test set (top) and the ${\sim}$70\% subset left after CAA/FairSteer reserve 30\% for training (bottom). TGS reserves no training data yet is net-positive in both blocks, while baselines are net-negative or only marginally positive. Net $\uparrow$ = errors fixed minus correct predictions broken; Fix\%/Break\% are rates. TGS (held-out): $\alpha$ and $\tau$ selected on the 30\% split (Section~\ref{sec:analysis}).}
\label{tab:baselines}
\end{table}

\subsection{TGS Transfers Across Models and Datasets}
\label{sec:generalization}

\textbf{TGS stays net-positive across three model families and four clinical and biomedical datasets, including a clinical task with a documented non-gender spurious feature and a biomedical QA task outside clinical notes.} Across models, we apply it to Mistral 7B Instruct ($\alpha{=}0.5$) and Qwen3 8B ($\alpha{=}2.0$) on i2b2 with the same $\tau=0.5$. Both are net-positive (Table~\ref{tab:generalization}). The higher $\alpha$ for Qwen3 reflects its larger logit scale.

On n2c2 2018, the i2b2 configuration transfers without retuning: TGS is net-positive and reduces gender-swap inconsistency from 37.2\% to 33.9\% (Appendix~\ref{sec:appendix_n2c2}).

SHAC ($\alpha{=}1.0$, $\tau{=}0.75$; Appendix~\ref{sec:appendix_grids}) tests a different spurious feature in clinical notes. Models are known to rely on alcohol and smoking terminology there rather than clinical content \citep{sakib2025-sdoh}. TGS achieves net +42 with a 78.9\% fix rate and reduces inconsistency with respect to these terms from 31.7\% to 16.8\% (Appendix~\ref{sec:appendix_inconsistency}). On a 3-class version of SHAC (negative / current / past), TGS also stays net-positive (net +29; 27.8\% fix, 5.8\% break; Appendix~\ref{sec:appendix_multiclass}).

Finally, beyond clinical notes, we evaluate on PubMedQA ($\alpha{=}1.0$, $\tau{=}0.75$; Appendix~\ref{sec:appendix_grids}), a biomedical research QA task with no counterfactual pairs. TGS achieves net +35 (34.5\% fix rate), so the correction does not require a known spurious feature. Table~\ref{tab:generalization} summarizes these results.

\begin{table}[t]
\centering
\small
\begin{tabular}{llrrr}
\toprule
Model & Dataset & Net & Fix\% & Break\% \\
\midrule
LLaMA & i2b2 & +201 & 17.7 & 1.3 \\
Mistral & i2b2 & +178 & 8.5 & 0.2 \\
Qwen3 & i2b2 & +414 & 22.3 & 0.2 \\
\midrule
LLaMA & SHAC & +42 & 78.9 & 2.7 \\
LLaMA & PubMedQA & +35 & 34.5 & 4.9 \\
LLaMA & n2c2 2018 & +51 & 6.6 & 1.2 \\
\bottomrule
\end{tabular}
\caption{Generalization across models and datasets. All results are net-positive. Top: cross-model on i2b2. Bottom: cross-dataset with LLaMA.}
\label{tab:generalization}
\end{table}

On i2b2, the same $\tau=0.5$ transfers across models without retuning; only $\alpha$ adjusts to each model's logit scale. TGS is net-positive on 41 of 48 model-disease combinations, with no disease consistently negative across models (Appendix~\ref{sec:appendix_perdisease}).

\subsection{TGS Is Robust to Its Hyperparameters and Prompt Wording}
\label{sec:analysis}

\textbf{On i2b2, every tested $(\alpha,\tau)$ configuration is net-positive, trajectory-based routing holds the break rate near zero, and the gains hold under a reworded prompt.} We vary each hyperparameter with the other fixed (Table~\ref{tab:sensitivity} in Appendix~\ref{sec:appendix_grids}): correction strength $\alpha$ trades off fix against break, with net gain peaking at $\alpha=0.5$, while across the conviction-growth threshold $\tau$ net gain stays in $[+184, +209]$. All tested i2b2 configurations remain net-positive, unlike baselines where small changes flip outcomes (prompt strategies span $-$6206 to $-$83). Grids for SHAC and PubMedQA are in Appendix~\ref{sec:appendix_grids}.

\paragraph{Held-out tuning.} The reported configurations follow the baselines' protocol. Selecting $\alpha$ and $\tau$ on a 30\% split and evaluating on the remaining 70\% leads to the same conclusions. On i2b2, it selects the reported configuration and yields the same net +146 on the CAA/FairSteer test subset (Table~\ref{tab:baselines}). On SHAC and PubMedQA, held-out tuning is net-positive in all 200 random splits (mean +28 and +21 on the held-out 70\%).

\paragraph{Prompt wording.} We rerun TGS on i2b2 with a paraphrased prompt that rewords every sentence while keeping the task and answer options (Appendix~\ref{sec:appendix_prompts}). With the same configuration, TGS remains net-positive (+234 vs.\ +201 with the original prompt) and reduces gender-swap inconsistency from 21.2\% to 16.1\%.

\paragraph{Routing protects correct predictions.} Removing trajectory-based routing and applying correction to all documents raises TP break from 0.1\% to 6.4\%, confirming that routing is essential for preserving correct predictions. A constant logit bias, which shifts every document's answer-token margin by the same amount, is net-negative at every tested strength and in both directions (Appendix~\ref{sec:appendix_output}). On i2b2, routing sends 47\% of documents to Reinforce, 13\% to Oppose, and 40\% to Leave unchanged; the break rate is 2.4\% for Reinforce and 0.3\% for Oppose (Appendix~\ref{sec:appendix_grids}).

\section{Related Work}
\label{sec:related}

\paragraph{Spurious correlations and shortcut learning.} Neural models frequently exploit surface-level patterns that correlate with labels but do not reflect the underlying task \citep{Geirhos_2020}, including syntactic heuristics in NLI \citep{mccoy-etal-2019-right} and shortcuts in fine-tuned and in-context LLMs \citep{tu-etal-2020-empirical, tang-etal-2023-large}. In clinical text, models may rely on demographic features or substance-related terms rather than clinical evidence \citep{Brown_2023,sakib2025-sdoh}. Training-time mitigations, including adversarial debiasing \citep{zhang2018mitigatingunwantedbiasesadversarial}, domain-invariant and invariant risk learning \citep{ganin2016domainadversarialtrainingneuralnetworks, arjovsky2020invariantriskminimization}, stable learning \citep{zhang2021deepstablelearningoutofdistribution}, and causal feature selection \citep{zhou2025fightingspuriouscorrelationstext}, require retraining, often with labels for the feature or for distinct environments. TGS leaves the model unchanged and corrects predictions at inference time.

\paragraph{Activation steering.} Representation engineering, CAA, and ITI identify directions for a concept in activation space and shift activations along them during inference \citep{zou2025representationengineeringtopdownapproach, rimsky-etal-2024-steering, li2024inferencetimeinterventionelicitingtruthful}. FairSteer applies correction only when a trained bias classifier fires \citep{li2025-fairsteer}, and \citet{ahsan2025elucidating} locate and patch gender information in LLMs used for healthcare. Each of these methods relies on external supervision (contrastive prompt pairs, labeled examples of the target behavior, or trained classifiers); TGS derives its correction signal from the model's own forward pass on the current input.

\paragraph{Logit lens and prediction trajectories.} The logit lens \citep{nostalgebraist2020logitlens} reads a model's token preferences at each layer, and the tuned lens \citep{belrose2023eliciting} refines this with learned per-layer affine probes, yielding intermediate predictions that are more predictive and reliable. Layer-wise predictions have also been used to change outputs: DoLa contrasts early and final layers at each decoding step \citep{chuang2024doladecodingcontrastinglayers}, early-exit methods stop once intermediate predictions agree \citep{zhou2020bertlosespatiencefast}, intermediate decoding recovers accuracy under misleading prompts \citep{halawi2024overthinkingtruthunderstandinglanguage}, and layer-wise commitment can modulate context injection \citep{fazli2026contextawaredecodingfaithfulvisionlanguage}. These methods act on every input; TGS uses trajectory shape to decide, per document, whether to correct and in which direction. Trajectories over decoding steps also guide sampling in masked diffusion models \citep{saini2026tabestrajectoryawarebackwardonentropysteering}; ours span the layers of an autoregressive model.

\paragraph{Logit-level correction.} Contextual calibration \citep{zhao2021calibrateuseimprovingfewshot} applies the same answer-token correction, estimated from a content-free input, to every input. TGS also adjusts only answer-token logits, but decides per document whether and in which direction, using the model's own trajectory (Appendix~\ref{sec:appendix_rw}).

\section{Conclusion}

We showed that a model's prediction trajectory, read with the logit lens, can guide selective inference-time correction for clinical classification. TGS decides from the model's own forward pass whether and in which direction to correct each prediction, with no counterfactual data, no knowledge of the spurious feature, and no retraining, as clinical settings require. Safe correction does not require distinguishing correct from incorrect predictions: trajectory shape indicates when a correction is safe and in which direction. On i2b2, TGS fixes 17.7\% of errors while breaking only 1.3\% of correct predictions. It is net-positive across three model families and four clinical and biomedical datasets, and its corrections concentrate on errors driven by the spurious feature, making predictions more consistent under gender swaps. Larger multi-class tasks and non-medical domains are natural next settings.

\section*{Limitations}
\label{sec:limitations}

This work has several limitations. First, the method requires a task with a fixed set of candidate answer tokens (e.g., Yes/No) over which the logit margin is computed, which covers classification and QA but not open-ended generation. Second, the hyperparameters $\alpha$ and $\tau$ must be selected per model-dataset combination, requiring a small grid search. Third, all experiments use 7--8B parameter instruction-tuned models in English, all on biomedical datasets; generalization to larger models, other languages, or non-biomedical domains remains untested. Evaluation on other medical QA datasets and on standard spurious-correlation benchmarks \citep{zhou2025fightingspuriouscorrelationstext} would test whether the same routing transfers beyond these settings. Fourth, the trajectory shapes that enable routing may not hold for all model architectures or tasks: TGS assumes that the logit lens provides informative intermediate predictions, which has been validated for transformer-based LLMs but not other architectures. Fifth, we test robustness to prompt wording with one paraphrase of the i2b2 prompt; broader prompt variation remains untested. Sixth, our evaluation covers two spurious features, gender (i2b2, n2c2 2018) and substance-use terminology (SHAC), but does not systematically span other demographic attributes such as race, age, or socioeconomic status, which we leave to future work.

% This work has several limitations. First, the method requires discrete answer tokens to compute logit margins, limiting it to classification and QA tasks; open-ended generation is not supported. Second, the correction strength $\alpha$ must be selected per model-dataset combination, requiring a small grid search; while $\tau$ is robust across its tested range, no single $\alpha$ is optimal across all settings. Third, all experiments use 7--8B parameter instruction-tuned models in English; generalization to larger models or other languages remains untested. Fourth, all datasets are in the medical domain; whether trajectory routing transfers to other domains is an open question. Fifth, the trajectory patterns that enable routing may not hold for all model architectures or tasks---TGS assumes that the logit lens provides informative intermediate predictions, which has been validated for transformer-based LLMs but not other architectures.

\section*{Ethics Statement}

All clinical experiments use de-identified patient data. The i2b2 2008 Obesity Challenge dataset, the n2c2 2018 cohort selection corpus, and the SHAC corpus (MIMIC-III subset) were accessed under institutional data use agreements, and all work was conducted on secure university servers following IRB protocols. PubMedQA contains only published biomedical abstracts with no patient data. Gender swapping is a diagnostic probe for reliance on spurious features and is not intended to capture the full spectrum of gender identity. TGS reduces but does not eliminate prediction errors driven by spurious features; we recommend further validation before any clinical deployment.

\bibliography{custom}
\appendix

\section{Steering Layer Selection}
\label{sec:appendix_steering}

The steering layer $\ell^*$ is selected per document by detecting where the model first commits to a decision. We compute binary entropy at each layer from the logit lens projection. Within the range $[10, L{-}4]$, we find the first layer $\ell'$ where the entropy decrease exceeds 50\% of the maximum observed drop, and set $\ell^* = \max(10,\; \ell' - 2)$, slightly before the detected transition. This adaptive selection enables the same procedure to work across model families without manual layer tuning.

Across models on i2b2, the selected $\ell^*$ values are: LLaMA 10--19 (median 14), Mistral 10--25 (median 14), Qwen3 10--29 (median 10). While these defaults were used for all experiments, varying the threshold from 30\% to 70\% and the offset from 1 to 3 produces only minor changes, so the method is robust to these choices.

\section{Prediction Inconsistency}
\label{sec:appendix_inconsistency}

Table~\ref{tab:inconsistency} reports the prediction inconsistency rate for LLaMA 3.1 8B Instruct on each i2b2 disease. Inconsistency is measured as the fraction of documents where the model's prediction changes when the gender token is swapped, with all other clinical content held constant.

\begin{table}[h]
\centering
\small
\begin{tabular}{lrrr}
\toprule
\textbf{Disease} & \textbf{Docs} & \textbf{Differ} & \textbf{\%} \\
\midrule
Hypertension & 531 & 284 & 53.5 \\
Heart Failure & 243 & 92 & 37.9 \\
Diabetes & 572 & 201 & 35.1 \\
ASCVD & 552 & 194 & 35.1 \\
Venous Insufficiency & 526 & 159 & 30.2 \\
Depression & 582 & 172 & 29.6 \\
Hypercholesterolemia & 503 & 139 & 27.6 \\
Obesity & 554 & 137 & 24.7 \\
Hypertriglyceridemia & 587 & 143 & 24.4 \\
PVD & 557 & 135 & 24.2 \\
GERD & 488 & 116 & 23.8 \\
OSA & 597 & 127 & 21.3 \\
Asthma & 572 & 80 & 14.0 \\
Osteoarthritis & 566 & 68 & 12.0 \\
Gallstones & 593 & 67 & 11.3 \\
Gout & 598 & 58 & 9.7 \\
\bottomrule
\end{tabular}
\caption{Prediction inconsistency rates across 16 i2b2 diseases. Each document is prompted under both male and female gender conditions; \textbf{Differ} counts cases where the prediction changes. Sorted by inconsistency rate.}
\label{tab:inconsistency}
\end{table}

Table~\ref{tab:inconsistency_tgs} reports the change in inconsistency after applying TGS ($\alpha{=}0.5$, $\tau{=}0.5$). TGS reduces inconsistency in 14 of 16 diseases, with the largest reductions on the most affected diseases. The two diseases where inconsistency increases (Hypertriglyceridemia, Venous Insufficiency) are also the two most net-negative in the fix/break evaluation.

\begin{table}[h]
\centering
\small
\resizebox{\columnwidth}{!}{%
\begin{tabular}{lrrrr}
\toprule
\textbf{Disease} & \textbf{Docs} & \textbf{Before \%} & \textbf{After \%} & \textbf{$\Delta$pp} \\
\midrule
Hypertension & 531 & 53.5 & 38.0 & $-$15.5 \\
Diabetes & 567 & 35.1 & 21.5 & $-$13.6 \\
Heart Failure & 243 & 37.9 & 28.0 & $-$9.9 \\
Hypercholesterolemia & 502 & 27.5 & 19.3 & $-$8.2 \\
Obesity & 553 & 24.6 & 17.2 & $-$7.4 \\
ASCVD & 548 & 34.9 & 30.1 & $-$4.8 \\
GERD & 487 & 23.8 & 19.9 & $-$3.9 \\
Gout & 596 & 9.4 & 7.4 & $-$2.0 \\
Depression & 582 & 29.6 & 27.8 & $-$1.8 \\
Osteoarthritis & 565 & 12.0 & 10.8 & $-$1.2 \\
Gallstones & 593 & 11.3 & 10.3 & $-$1.0 \\
PVD & 556 & 24.3 & 23.9 & $-$0.4 \\
Asthma & 572 & 14.0 & 13.8 & $-$0.2 \\
OSA & 590 & 20.8 & 20.7 & $-$0.1 \\
Hypertriglyceridemia & 587 & 24.4 & 27.8 & +3.4 \\
Venous Insufficiency & 526 & 30.2 & 33.8 & +3.6 \\
\midrule
\textbf{Overall} & \textbf{8598} & \textbf{25.1} & \textbf{21.5} & $\mathbf{-3.6}$ \\
\bottomrule
\end{tabular}%
}
\caption{Prediction inconsistency before and after TGS. Sorted by improvement ($\Delta$pp). TGS reduces inconsistency in 14/16 diseases, with an overall reduction from 25.1\% to 21.5\%.}
\label{tab:inconsistency_tgs}
\end{table}

\paragraph{Which errors TGS corrects.} An error of the original prediction is gender-dependent if the other gender version of the same document is predicted correctly. TGS fixes 19.8\% of gender-dependent errors but only 4.9\% of errors that persist under both genders. Gender-dependent errors might simply be easier to flip, since the model already answers correctly for the other gender. To test this, we add a uniform logit bias toward the correct answer for every error, applied in the same way as the TGS correction (Section~\ref{sec:generation}). This uniform push corrects both kinds of error at nearly the same rate (13.7\% vs.\ 14.8\%).

On SHAC, inconsistency with respect to alcohol and smoking mentions (how often removing them from a note changes the prediction) falls from 31.7\% to 16.8\% after TGS.

\section{Per-Disease Results}
\label{sec:appendix_perdisease}

Table~\ref{tab:perdisease} reports the full per-disease breakdown for LLaMA on i2b2 ($\alpha{=}0.5$, $\tau{=}0.5$). Performance scales with disease error prevalence: diseases with many errors in the original predictions (e.g., Diabetes, Hypertension) show the largest net gains, while diseases with few errors yield smaller or negative net outcomes. Across all three models, TGS is net-positive on 41 of 48 model-disease combinations.

\begin{table}[h]
\centering
\small
\begin{tabular}{lrrr}
\toprule
\textbf{Disease} & \textbf{Net} & \textbf{Fix\%} & \textbf{Break\%} \\
\midrule
Diabetes & +77 & 39.3 & 0.2 \\
Hypertension & +75 & 27.3 & 0.7 \\
Obesity & +41 & 30.9 & 0.2 \\
Hypercholesterolemia & +39 & 28.2 & 0.6 \\
Heart Failure & +22 & 27.5 & 0.0 \\
GERD & +18 & 14.5 & 0.1 \\
Gout & +12 & 21.0 & 0.1 \\
Gallstones & +10 & 16.0 & 0.3 \\
Depression & +6 & 6.9 & 0.5 \\
Osteoarthritis & +6 & 8.4 & 0.2 \\
Asthma & +1 & 10.9 & 0.9 \\
OSA & $-$5 & 14.9 & 1.6 \\
PVD & $-$9 & 9.3 & 2.1 \\
ASCVD & $-$20 & 5.1 & 3.4 \\
Hypertriglyceridemia & $-$33 & 4.2 & 4.0 \\
Venous Insufficiency & $-$39 & 4.0 & 5.4 \\
\bottomrule
\end{tabular}
\caption{Per-disease results for TGS on i2b2 (LLaMA, $\alpha{=}0.5$, $\tau{=}0.5$). Fix\% and Break\% are computed over all error types and all correct predictions, respectively.}
\label{tab:perdisease}
\end{table}

\section{Hyperparameter Grids}
\label{sec:appendix_grids}

Table~\ref{tab:sensitivity} reports the i2b2 sensitivity analysis, and Tables~\ref{tab:grid_shac} and~\ref{tab:grid_pubmedqa} report the full hyperparameter sweep for SHAC and PubMedQA. On PubMedQA, all 16 configurations are net-positive. On SHAC, every configuration with $\alpha \geq 0.75$ is net-positive (+21 to +42), while $\alpha \leq 0.50$ is net-negative at every $\tau$.

\begin{table}[h]
\centering
\small
\begin{tabular}{crrr}
\toprule
& Net & Fix\% & Break\% \\
\midrule
\multicolumn{4}{l}{\textit{$\alpha$ sensitivity ($\tau=0.5$ fixed)}} \\
$\alpha=0.25$ & +114 & 8.6 & 0.5 \\
$\alpha=0.50$ & \textbf{+201} & 17.7 & 1.3 \\
$\alpha=0.75$ & +174 & 24.5 & 2.5 \\
$\alpha=1.00$ & +21 & 29.7 & 4.3 \\
\midrule
\multicolumn{4}{l}{\textit{$\tau$ sensitivity ($\alpha=0.5$ fixed)}} \\
$\tau=0.25$ & +209 & 18.1 & 1.3 \\
$\tau=0.50$ & +201 & 17.7 & 1.3 \\
$\tau=0.75$ & +199 & 17.4 & 1.2 \\
$\tau=1.00$ & +184 & 16.4 & 1.2 \\
\bottomrule
\end{tabular}
\caption{Sensitivity analysis (LLaMA, i2b2, 16 diseases). $\alpha$ controls the fix--break tradeoff; performance is robust to $\tau$ within its tested range.}
\label{tab:sensitivity}
\end{table}

\begin{table}[h]
\centering
\small
\begin{tabular}{crrrr}
\toprule
& \multicolumn{4}{c}{$\tau$} \\
\cmidrule(l){2-5}
$\alpha$ & 0.25 & 0.50 & 0.75 & 1.00 \\
\midrule
0.25 & $-$32 & $-$32 & $-$32 & $-$32 \\
0.50 & $-$3 & $-$3 & $-$3 & $-$3 \\
0.75 & +21 & +33 & +40 & +40 \\
1.00 & +21 & +33 & \textbf{+42} & +40 \\
\bottomrule
\end{tabular}
\caption{Net gain across $\alpha \times \tau$ configurations on SHAC (LLaMA). Bold: reported configuration ($\alpha{=}1.0$, $\tau{=}0.75$).}
\label{tab:grid_shac}
\end{table}

\begin{table}[h]
\centering
\small
\begin{tabular}{crrrr}
\toprule
& \multicolumn{4}{c}{$\tau$} \\
\cmidrule(l){2-5}
$\alpha$ & 0.25 & 0.50 & 0.75 & 1.00 \\
\midrule
0.25 & +6 & +6 & +6 & +6 \\
0.50 & +17 & +17 & +17 & +17 \\
0.75 & +26 & +30 & +32 & +32 \\
1.00 & +26 & +29 & \textbf{+35} & +33 \\
\bottomrule
\end{tabular}
\caption{Net gain across $\alpha \times \tau$ configurations on PubMedQA (LLaMA, 890 documents). All configurations are net-positive. Bold: reported configuration ($\alpha{=}1.0$, $\tau{=}0.75$).}
\label{tab:grid_pubmedqa}
\end{table}

\paragraph{Routing distribution.} For analysis, we group documents by which branch of Eq.~\ref{eq:routing} they take: \textsc{Reinforce} ($e_d = +m_{\mathrm{e}}$), \textsc{Oppose} ($e_d = -\max(m_{\mathrm{e}},0)$), and \textsc{Leave unchanged} ($e_d = 0$). On i2b2, 47\% of documents are routed to \textsc{Reinforce}, 13\% to \textsc{Oppose}, and 40\% to \textsc{Leave unchanged}. On PubMedQA, the distribution is 67\%/12\%/21\%. Per-action break rates confirm the routing's safety ordering: \textsc{Reinforce}, the most aggressive action, has a 2.4\% break rate on i2b2; \textsc{Oppose}, designed as a conservative correction, has 0.3\%; \textsc{Leave unchanged} changes no predictions by design.

\section{n2c2 2018 Cohort Selection}
\label{sec:appendix_n2c2}

The corpus contains 288 patients annotated for 13 trial-eligibility criteria. We ask one Yes/No question per patient and criterion, using the i2b2 prompt with the criterion definition in place of the disease, and apply TGS with the i2b2 configuration ($\alpha{=}0.5$, $\tau{=}0.5$). TGS is net-positive (+51; 6.6\% fix, 1.2\% break) and reduces gender-swap inconsistency on 10 of 13 criteria (37.2\% $\rightarrow$ 33.9\%).

\section{Output-Level Controls}
\label{sec:appendix_output}

We test two corrections that act on the answer tokens of every document in the same way, without trajectory information. Both use the same i2b2 documents and scoring rule as TGS (Section~\ref{sec:evaluation}).

\paragraph{Constrained decoding.} We also run constrained decoding, which forces the model to answer \texttt{Yes} or \texttt{No}. It leaves all existing Yes/No predictions unchanged and changes only outputs that were neither Yes nor No.

\paragraph{Constant logit bias.} We add the same bias to the answer-token logits of every document, applied in the same way as the TGS correction (Section~\ref{sec:generation}), toward \texttt{Yes} ($+$) or \texttt{No} ($-$). Table~\ref{tab:constant_bias} shows that every nonzero strength is net-negative. Selecting the strength on the full test set or on the 30\% split chooses no bias.

\begin{table}[h]
\centering
\small
\begin{tabular}{rrrr}
\toprule
Bias & Fix & Break & Net \\
\midrule
$+$0.10 & 219 & 274 & $-$55 \\
$-$0.10 & 80 & 116 & $-$36 \\
$+$0.25 & 431 & 580 & $-$149 \\
$-$0.25 & 307 & 430 & $-$123 \\
$+$0.50 & 798 & 1382 & $-$584 \\
$-$0.50 & 505 & 842 & $-$337 \\
$+$1.00 & 1198 & 4015 & $-$2817 \\
$-$1.00 & 706 & 1581 & $-$875 \\
\bottomrule
\end{tabular}
\caption{Constant logit bias on i2b2 (LLaMA). TGS: net +201.}
\label{tab:constant_bias}
\end{table}

\section{Multi-Class Extension}
\label{sec:appendix_multiclass}

The core TGS quantities generalize to $K > 2$: the margin becomes the gap between the top and runner-up tokens, agreement becomes whether the same token leads at both early and late layers, and conviction growth becomes the widening of this gap. The logit bias (Eq.~\ref{eq:bias}) adjusts the top two competing tokens accordingly.

As initial evidence that TGS extends past binary decisions, we evaluate this formulation on a 3-class version of SHAC (negative / current / past), obtained by splitting the positive class into \emph{current} and \emph{past} ($\alpha{=}1.5$, $\tau{=}0.5$). TGS remains net-positive (net +29; 27.8\% fix, 5.8\% break). A broader multi-class study remains future work.

\section{Extended Related Work}
\label{sec:appendix_rw}

This section gives more detail on the methods discussed in Section~\ref{sec:related}.

\paragraph{Activation steering.} Representation engineering \citep{zou2025representationengineeringtopdownapproach} studies how high-level concepts can be identified and manipulated in a model's activation space, enabling targeted behavioral control. Contrastive Activation Addition \citep{rimsky-etal-2024-steering} operationalizes this by computing steering vectors from paired prompts that differ in the target behavior, then adding these vectors uniformly during generation. Inference-Time Intervention \citep{li2024inferencetimeinterventionelicitingtruthful} takes a similar approach for truthfulness, identifying intervention directions across attention heads whose activations are predictive of truthful responses. FairSteer \citep{li2025-fairsteer} adds selectivity via a trained classifier that detects bias signatures and applies correction only when triggered. In healthcare, \citet{ahsan2025elucidating} show that gender information is localized in specific MLP layers of open LLMs and can be manipulated at inference time via patching.

\paragraph{Logit lens and prediction trajectories.} The logit lens \citep{nostalgebraist2020logitlens} projects intermediate hidden states through the unembedding matrix to reveal a model's evolving token preferences across layers; the tuned lens \citep{belrose2023eliciting} learns a per-layer affine probe for the same readout. DoLa \citep{chuang2024doladecodingcontrastinglayers} contrasts final- and early-layer distributions at each decoding step to improve factuality, early-exit methods stop computation once intermediate predictions agree \citep{zhou2020bertlosespatiencefast}, decoding from an intermediate layer can recover accuracy lost in later layers under misleading prompts \citep{halawi2024overthinkingtruthunderstandinglanguage}, and \citet{fazli2026contextawaredecodingfaithfulvisionlanguage} use layer-wise commitment to modulate context injection during decoding.

\paragraph{Logit-level correction.} Contextual calibration \citep{zhao2021calibrateuseimprovingfewshot} reduces a model's preference for particular labels by rescaling answer-token probabilities with a correction estimated from a content-free input, applied identically to every input and requiring no counterfactual data. Our constant-bias control (Appendix~\ref{sec:appendix_output}) applies one such fixed shift to every document and is net-negative at every tested strength (Table~\ref{tab:constant_bias}).

\section{Prompt Templates}
\label{sec:appendix_prompts}

All experiments use the LLaMA 3.1 chat template with a system message and user message. Below are the templates for each dataset.

\paragraph{i2b2.}
\begin{quote}
\small
\textbf{System:} You are a medical expert analyzing patient records. \\[4pt]
\textbf{User:} Review this patient record: \\
\texttt{[patient text]} \\
The patient is a \texttt{[gender]}. Based on both explicit mentions AND implicit clinical indicators, determine if the patient has \texttt{[disease]}. \\
Your task is to determine one of these judgments: \\
-- Yes: if there is clear evidence the patient has \texttt{[disease]} \\
-- No: if there is no evidence or evidence against \texttt{[disease]} \\
-- Questionable: if there are some indicators but insufficient evidence \\
Respond with EXACTLY one of these words: Yes, No, or Questionable.
\end{quote}

\paragraph{Paraphrased i2b2 prompt (Section~\ref{sec:analysis}).}
\begin{quote}
\small
\textbf{System:} You are a physician reviewing patient charts. \\[4pt]
\textbf{User:} Examine this patient record: \\
\texttt{[patient text]} \\
The patient is a \texttt{[gender]}. Considering both explicit documentation and implicit clinical evidence, determine whether the patient has \texttt{[disease]}. \\
Select one of these judgments: \\
-- Yes: there is clear evidence that the patient has \texttt{[disease]} \\
-- No: there is no evidence of \texttt{[disease]}, or there is evidence against it \\
-- Questionable: there are some indicators, but not enough evidence \\
Respond with EXACTLY one of these words: Yes, No, or Questionable.
\end{quote}
With this prompt, TGS fixes 35.0\% of errors and breaks 3.2\% of correct predictions.

\paragraph{SHAC.}
\begin{quote}
\small
\textbf{System:} You are an AI assistant specialized in extracting and analyzing social history information from medical notes. \\[4pt]
\textbf{User:} Choose the best StatusTime value for the $\langle$Drug Use$\rangle$ (Drug) from the note by selecting one of the options below: \\
(a)~none. (b)~current. (c)~past. (d)~not applicable. \\
IMPORTANT: Respond with ONLY the letter and label of your choice, e.g., `(a) none.' Do not include any other text. \\
Note: \texttt{[clinical note]}
\end{quote}

\paragraph{PubMedQA.}
\begin{quote}
\small
\textbf{System:} You are a biomedical research expert. Your task is to answer research questions based on evidence from scientific abstracts. \\[4pt]
\textbf{User:} Research Question: \texttt{[question]} \\
Abstract: \texttt{[abstract text]} \\
Based on the abstract, does the evidence support the research question? Respond with EXACTLY one word: Yes or No.
\end{quote}

\subsection{Prompt-Based Baselines}
\label{sec:appendix_baseline_prompts}

The three prompt baselines modify the i2b2 prompt as follows. The user message remains unchanged unless noted.

\paragraph{System-clinical.} Replaces the system message:
\begin{quote}
\small
You are a medical expert analyzing patient records. Base your assessment strictly on documented clinical findings, diagnoses, medications, laboratory results, and procedures.
\end{quote}

\paragraph{System-objective.} Replaces the system message:
\begin{quote}
\small
You are a medical expert analyzing patient records. Be objective and evidence-based. Your assessment should rely solely on the clinical information documented in the record.
\end{quote}

\paragraph{Chain-of-thought.} Replaces the final response instruction with:
\begin{quote}
\small
Before giving your final answer, reason step by step about the clinical evidence for and against this diagnosis. \\[4pt]
After your reasoning, provide your final answer on a new line in the format: \\
Final Answer: Yes, No, or Questionable
\end{quote}

\section{Computational Resources}
\label{sec:appendix_compute}

No base-LLM parameters are updated in any experiment. TGS and the prompt baselines are inference-only; CAA estimates steering vectors from its training subset, and FairSteer additionally trains per-layer linear probes. We evaluate three instruction-tuned models (LLaMA 3.1 8B Instruct, Mistral 7B Instruct v0.3, and Qwen3 8B), loaded in bfloat16 on two NVIDIA A100 80GB GPUs. TGS adds one forward pass per document to extract the logit-lens trajectory; documents receiving a nonzero correction additionally undergo a modified generation pass. The full set of experiments reported in this paper, including the baseline sweeps and the hyperparameter grids in Appendix~\ref{sec:appendix_grids}, required approximately 100 GPU hours.

\end{document}